\documentclass[runningheads]{llncs}
\usepackage{lineno}
\usepackage{graphicx}
\usepackage{cite}
\usepackage{amsmath,amssymb,amsfonts}
\usepackage{algorithmic}
\usepackage{algorithm}
\usepackage{graphicx}
\usepackage{textcomp}
\usepackage{booktabs}
\usepackage{xcolor}
\usepackage{comment}
\usepackage{array}

\begin{document}
\pdfoutput=1
\title{Unsupervised Deep Features for Privacy \\Image Classification
}

\author{Chiranjibi Sitaula
\and
Yong Xiang
\and
Sunil Aryal
\and
Xuequan Lu
}

\authorrunning{C. Sitaula et al.}
%

\institute{School of Information Technology, Deakin University, Australia\\ 
\email{\{csitaul, yong.xiang, sunil.aryal, xuequan.lu\}@deakin.edu.au} }


%
\maketitle              
\begin{abstract}

Sharing images online poses security threats to a wide range of users due to the unawareness of privacy information. Deep features have been demonstrated to be a powerful representation for images. However, deep features usually suffer from the issues of a large size and requiring a huge amount of data for fine-tuning. In contrast to normal images (e.g., scene images), privacy images are often limited because of sensitive information. In this paper, we propose a novel approach that can work on limited data and generate deep features of smaller size. For training images, we first extract the initial deep features from the pre-trained model and then employ the K-means clustering algorithm to learn the centroids of these initial deep features.
We use the learned centroids from training features to extract the final features for each testing image and encode our final features with the triangle encoding. 
 To improve the discriminability of the features, we further perform the fusion of two proposed unsupervised deep features obtained from different layers.  Experimental results show that the proposed features outperform state-of-the-art deep features, in terms of both classification accuracy and testing time.

\end{abstract}
\section{Introduction}
Privacy image classification is becoming increasingly important nowadays, owing to the prevalent presence of social media on the web where people share personal and private images. The privacy image classification systems allow people to know if the images they share are private or public.
Private images, such as images involving families, usually involve private information about the users. By contrast, public images generally involve scenes, objects, animals and so on, and do not include private information. The purpose of the privacy image classification is to make people alert while sharing images online. People sometimes may be unaware of whether they are doing right or wrong when sharing their images. In such cases, a system that is capable of classifying private and public images is very useful to users.

\begin{figure}[tb]
\begin{center}
\includegraphics[width=\textwidth,height=4.2cm,keepaspectratio]{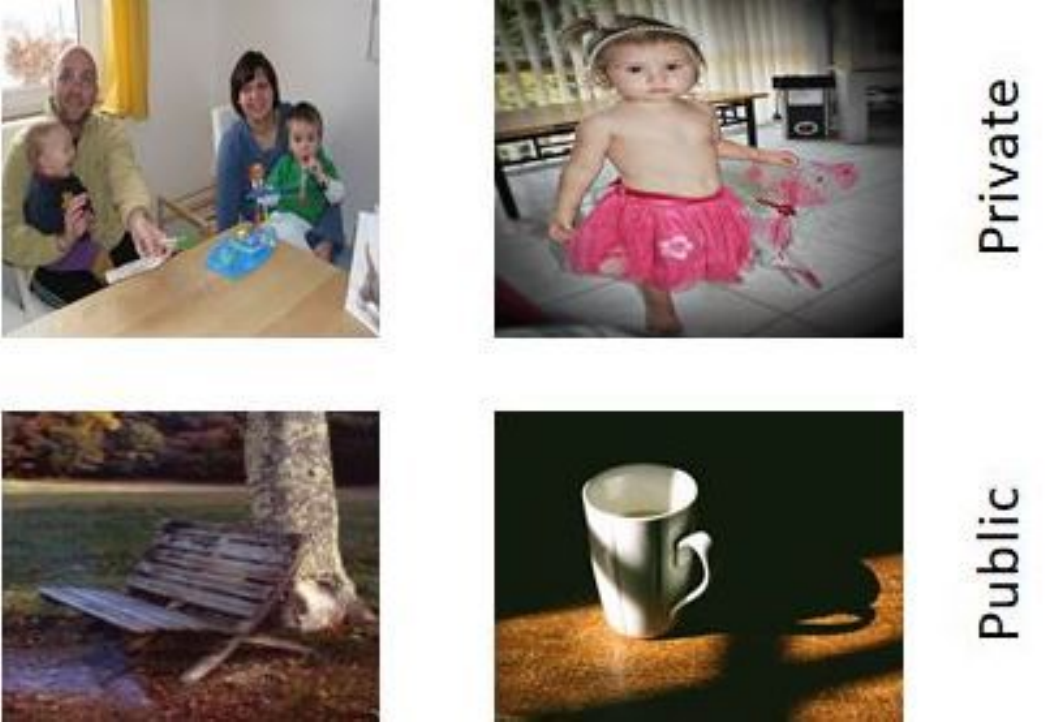}
\end{center}
\caption{Images showing the private and public images from PicAlert\cite{zerr2012privacy} dataset.} 
\label{fig:0}
\end{figure}

For image classification, feature extraction from images is a fundamental step. 
Privacy images are challenging for classification, because they may contain high within-class dissimilarity. As shown in Fig. \ref{fig:0}, we observe in both categories (private and public) that they have such patterns. Fortunately, there are only two categories in privacy images so that we do not need to consider such varying patterns as in other scene image classification which have far more than two categories \cite{8746125}.

In general, the existing feature extraction methods for privacy images comprise of traditional vision-based methods\cite{zerr2012privacy}, deep learning-based methods\cite{tonge2015privacy,tonge2016image,zhong2017group,tonge2018use,tran2016privacy,tonge2018uncovering}, and semantic approaches \cite{spyromitros2016personalized,sitaula2019tagbased}. While comparing traditional vision-based features against the deep learning-based features, we notice a significant improvement in classification accuracy with the aid of the latter features learned from the pre-trained deep learning models. By the help of the fine-tuned deep learning models, it can even achieve a higher classification accuracy which required a massive amount of data \cite{tran2016privacy}. Nevertheless, in the task of privacy image classification, there is a very limited amount of data due to privacy issues. Simply extracting features from intermediate layers of those models makes the size of the features higher, thereby increasing computational burden during classification. 
To sum up, these existing methods on privacy images suffer from \textbf{two} problems:  1) the curse of dimensionality of features; and 2) requirements of massive data if we want to obtain a fine-tuned model or new deep learning model.
As such, feature extraction methods favoring a low feature size and limited data are particularly needed for the task of privacy image classification.

In this paper, we propose a novel approach to extract the features of privacy images with the assistance of unsupervised feature learning, which not only works on a limited amount of privacy images but also yields a lower feature size. Inspired by the work in \cite{tonge2018use}, where the authors 
claim the efficacy of the pre-trained models against the fine-tuned models over privacy images, we also choose a pre-trained model in this work. Specifically, among several pre-trained models, we choose the ResNet-50\cite{he2016deep} model,  
which has been found to have a lower error rate for the classification of different types of images than the state-of-the-art deep learning models such as VGG-Net\cite{simonyan2014very} and GoogleNet\cite{szegedy2015going}. Furthermore, the ResNet-50 also has a lower number of layers than its other versions (ResNet-101 and ResNet-152), thereby having a faster speed. 
To perform unsupervised feature learning, we perform the K-means clustering on the deep features extracted from the ResNet-50\cite{he2016deep} which has been pre-trained with a large dataset of labeled images (i.e., ImageNet\cite{imagenet_cvpr09}). Then, we encode the features using the triangle encoding\cite{coates2011analysis} to achieve our unsupervised deep features. 
The K-means clustering can yield centroids of patterns (contexts) for privacy images. The features of the clustering method are (1) discriminable patterns of privacy images and (2) a lower feature size due to its dimension reduction capability.
We tested our unsupervised deep features on PicAlert\cite{zerr2012picalert} and found that our features can produce better classification accuracies than deep learning features extracted by state-of-the-art models.

\section{Related works}

Several studies have explored the privacy image classification problem with the use of different types of features such as SIFT (Scale Invariant Feature Transform) and RGB (Red Green Blue)~\cite{zerr2012privacy}, textual and deep learning based features ~\cite{tonge2015privacy,tonge2016image,zhong2017group,tonge2018use,tran2016privacy,tonge2018uncovering}, semantic features~\cite{spyromitros2016personalized}, and so on.

Zerr et al. \cite{zerr2012picalert}
used various types of visual features such as quantized SIFT, color histogram, brightness and sharpness and the text features of the image. They have shown that the features designed by the fusion of textual and visual features are prominent than the visual features only. Similarly, the authors 
 in~\cite{tonge2015privacy,tonge2016image,tonge2018use} emphasized the usage of textual features such as deep tags (object tags and scene tags) and user tags (user annotated tags) 
 based features for the classification of privacy images and claimed that the features designed based on tags outperform the state-of-the-art features such as SIFT, GIST (Generalized Search Tree) and fully connected features (${FC}$-features of VGG-Net). Zhong et al. \cite{zhong2017group} chose $FC$-features of a deep learning model for the group-based personalized approach which further proved the applicability of high-level features such as $FC$-features for this domain.
Similarly, Spyromitros et al. \cite{spyromitros2016personalized} 
explored the semantic features based on the output of a large array of classifiers. Their proposed semantic features outperform the generic traditional vision-based features such as SIFT, EDCH (Edge Direction Coherence) feature, etc.

More recently, 
Tonge et al. \cite{tonge2018uncovering}
explored textual features based on the pre-trained deep learning model, which yielded the scene information of the image, called scene tags. The authors unveiled that the combination of such scene tags with user tags and object tags outperforms features of individual tags. 
Likewise, Tran et al. \cite{tran2016privacy} 
extracted hierarchical features by the concatenation of object features and convolutional features. For the experiments, the authors used two pipelined CNNs (Convolutional Neural Networks). The $FC$-features obtained after the fine-tuning operation over two deep learning models were concatenated to get the final hierarchical features of the image. 
Their method requires a massive amount of images for training.
However, in the recent research by  Tonge et al. \cite{tonge2018use} 
the features extracted from the pre-trained model (${FC}$-features of AlexNet\cite{krizhevsky2012imagenet}) outperform the hierarchical features 
extracted from the fine-tuned deep learning models \cite{tran2016privacy}. 
Thus, 
task-generic features which are extracted from the pre-trained models, became more prominent than task-specific features which are extracted from fine-tuned deep learning models, for privacy images.
This opens a door to take advantage of the pre-trained models for the feature extraction of privacy images, given a limited amount of training images.

\section{Unsupervised Features Extraction}

\begin{figure*}[b]
\begin{center}
\includegraphics[width=\textwidth, keepaspectratio]{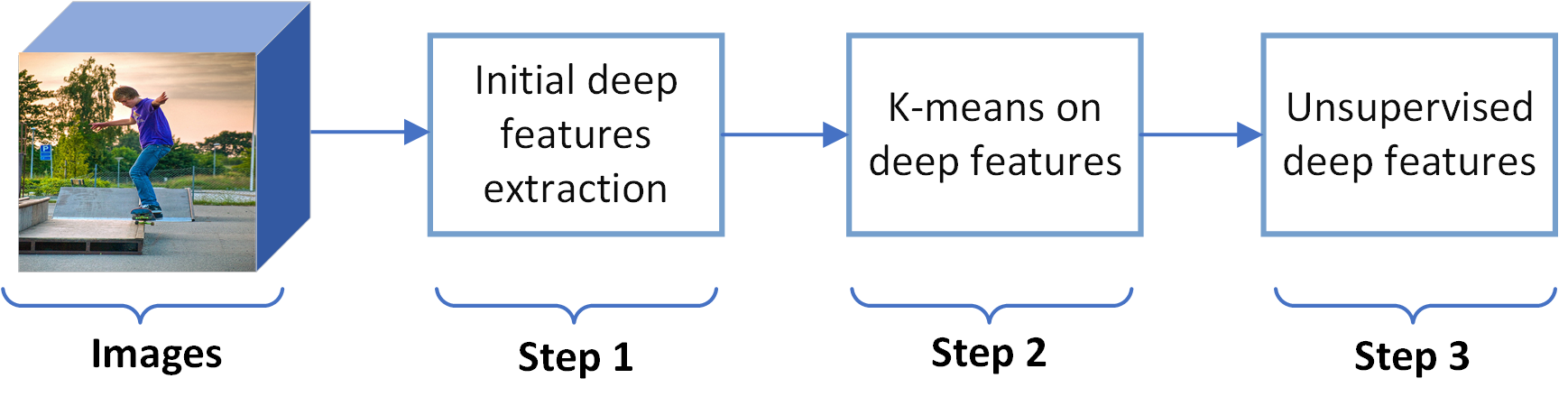}
\end{center}
\caption{Block diagram of the extraction of our proposed unsupervised deep features (UDF) encoding.}
\label{fig:1}
\end{figure*}
To extract the  unsupervised deep features, we chose the pre-trained 
ResNet-50 model. A pre-trained model is favorable owing to the following reasons: 1) fine-tuned models require massive data to overcome overfitting, and 2) there is a very limited amount of private images for the study. 
The overall approach, shown as a block diagram in Fig.~\ref{fig:1}, consists of three main steps to extract the unsupervised deep features, namely: initial deep features extraction (Sec. 3.1), K-means clustering on deep features (Sec. 3.2), and unsupervised deep features encoding (Sec. 3.3). 

\subsection{Initial deep features extraction}

\begin{figure*}[b]
\begin{center}
\includegraphics[width=\textwidth, height=3.6cm,keepaspectratio]{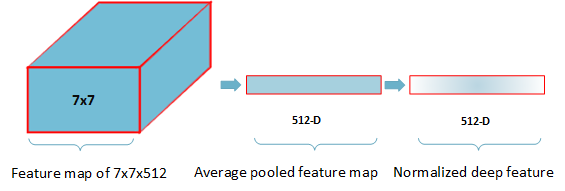}
\end{center}
\caption{The steps to extract the initial deep features of the selected activation layers (e.g., activation 48) from the pre-trained ResNet-50 model.} 
\label{fig:2}
\end{figure*}

We take the features from the top activation layers 
as the candidate deep features which
can better represent images based on the objects' details in the images\cite{8746125}.
The original dimension of the deep features from the activation layers is $7*7*512$, which provides 
512-D features (each feature map is $7*7$). 
To represent a feature map as a single value, we operate the global average pooling that exploits the properties of deep features with both high and low values. This results in a 512-D vector of an image where each component represents its corresponding feature map.
Let $H$, $W$, and $D$ denote the height, width 
, and depth of the candidate deep features of the top activation layers of the ResNet-50 model. 
\begin{equation}
{f(x_a)}=\frac{1}{H*W}*\{\sum_{i=1}^{H*W}{x_{a1}^i}, \sum_{i=1}^{H*W}{x_{a2}^i}, \sum_{i=1}^{H*W}{x_{a3}^i},...\sum_{i=1}^{H*W}{x_{aD}^i}\},
\label{eq:1}
\end{equation}
where $f(x_a)$ 
is the average pooled features of image $x_a$ based on the feature maps $\{x_{a1}^{i}, x_{a2}^{i}, \cdots, x_{aD}^{i}\}_{i=1}^{H*W}$. Eq.~\eqref{eq:1} computes the representative values of the corresponding feature maps.

The pooled features obtained from Eq.~\eqref{eq:1} are further processed by the two normalization strategies: power-normalization and L2-normalization. We first use the signed square root norm of the features for power-normalization and then perform L2-normalization, due to their higher performance\cite{lin2017bilinear,lin2015bilinear}.
\begin{equation}
{f(x'_a)}=\sqrt{{f(x_a)}}
\label{eq:2}
\end{equation}
Eq.~\eqref{eq:2} calculates the square root based power normalization ($f(x'_a)$) of each element of the average pooled feature vector $f(x_a$). Now, the features are normalized, as shown in the Eq.~\eqref{eq:3}.

\begin{equation}
{f(x''_a)}=\frac{f(x'_a)}{\| f(x'_a)\|_2}
\label{eq:3}
\end{equation}

Similarly,  Eq.~\eqref{eq:3} yields $f(x''_a)$, which is the L2-normalization of each element of the feature vector $f(x'_a)$.
The feature vectors of images extracted from Eq.~\eqref{eq:3} will be used to perform K-means clustering to learn the centroids (Sec. 3.2). 

Table~\ref{tab:1} lists detailed information about the layers used in this work. The first five activation layers are 512-D with a feature map size of $7*7$. 
For the average pooling layer (avg\_pool), the dimension is 2048-D in the ResNet-50 model with a feature map size of $1*1$. 
We perform global averaged pooling of each feature map to get the aggregated value of the corresponding feature map.
\begin{table}[tb]
\centering
\caption{Deep layers with sizes of feature maps and features from the pre-trained ResNet-50 model. The names in the bracket represent the activation layer name of the ResNet-50 model. We call these layers such as 42, 44, 45 and so on as methods because they output features. }
\begin{center}
\label{tab:1}
\begin{tabular}{m{5cm} m{2cm} m{2.3cm}}
\hline
\textbf{Methods} &  \textbf{Feat. map} & \textbf{Feat. size}\\
\hline 
\textbf{ResNet-50(42)} & ${7*7}$ & 512-D \\
\textbf{ResNet-50(44)} & ${7*7}$ & 512-D \\
\textbf{ResNet-50(45)} & ${7*7}$ & 512-D\\
\textbf{ResNet-50(47)} & ${7*7}$ & 512-D\\
\textbf{ResNet-50(48)} & ${7*7}$ & 512-D \\
\textbf{ResNet-50(avg\_pool)} & ${1*1}$ & 2048-D \\
\hline
\end{tabular}
\end{center}
\end{table}

\subsection{K-means clustering over deep features}  

We perform K-means clustering to learn the centroids of the initial deep features for the training dataset. Firstly, we set $k$ as an initial centroid number.
Let $c^{k}$ represent the $k^{th}$ cluster center. The $k$ clusters and centroids are optimized based on the distances of data points
to centroids. $k$ is set to $250$ (Sec. \ref{k_analysis}) which empirically produces a higher accuracy than others. 
While there are more delicately designed clustering algorithms, K-means is easy and simple to use, and we found it is effective in our context.

\subsection{Unsupervised deep features encoding}
After the calculation of the learned centroids $\{c^{k}\}$, we calculate the strength of all the initial deep features using the triangle encoding technique\cite{coates2011analysis}
which has a higher performance than hard assignment coding schemes as described by Coates et al.~\cite{coates2011analysis}. 
\begin{equation}
{f(\hat{x_a})}=max\{0, {\mu-{z_k}\}},
\label{eq:4}
\end{equation}
\newcommand{\norm}[1]{\left\lVert#1\right\rVert}
where ${z_k}$=
$d{(f(x''_a),c^k)}$
and $\mu$ is the average distance of all $f(x''_a)$ to all centriods and $f(\hat{x_a})$ denotes the unsupervised deep features in Eq.~\eqref{eq:4}.
\begin{equation}
d{(f(x''_a),c^k)}=\sqrt{(\sum{(f(x''_a)-c^k})^2}
\label{eq:5}
\end{equation}
We calculate the Euclidean distances between any two points, shown in Eq.~\eqref{eq:5}. 
After calculating the average distances from the corresponding initial features, we need to check if one distance is below or above its corresponding average distance. This is because the distances to all the centroids reveal the implicit relationship among centroids for the corresponding initial deep features.  To do so, we set 
the distance to $0$ if the distance is above the average distance. Otherwise we set it as the
difference between the average distance and Euclidean distance of the corresponding point.
Through this scheme, we are able to identify the importance of corresponding initial deep features to all centroids, which further facilitates the encoding of the features.
 In this work, the dimension of the resulting unsupervised deep features are $k$. Here, $k=250$ resulting in a 250-D vector for each privacy image.
\begin{algorithm}[t]
 \caption{Unsupervised deep features of training images}
 \begin{algorithmic}[1]
 \renewcommand{\algorithmicrequire}{\textbf{Input:}}
 \renewcommand{\algorithmicensure}{\textbf{Output:}}
 \REQUIRE $f(x'')\leftarrow$training initial deep features, $k\leftarrow$number of cluster centroids
 \ENSURE  $f(\hat{x})\leftarrow$training unsupervised deep features,\\ $c^k\leftarrow$cluster centroids of training features
 \STATE Perform K-means clustering on $f(x'')$ and extract $c^k$ centroids.
  \FOR {$i = 0$ to $n$}
   \FOR {$j = 0$ to $k$}
  \STATE $\sum$=$\sum_j$d($f(x''_i)$, $c^j$) 
  \ENDFOR
  \FOR {$l = 0$ to $k$}
\STATE ${\mu\leftarrow} \sum/k$
\STATE ${z_l}\leftarrow d(f(x''_i),c^l)$
 \STATE $\hat{x}_l\leftarrow  max\{0, \mu-z_l\}$
  \ENDFOR
  \STATE $f(\hat{x_i}) \leftarrow \hat{x_l}$
    \ENDFOR
 \RETURN $f(\hat{x})$
 \end{algorithmic}
 \label{algo:0} 
 \end{algorithm}

We assume that the initial deep features are represented by $f(x'')$ in Alg. \ref{algo:0} for training. To extract the proposed features, we perform several steps. First of all, we perform K-means clustering over such deep features to obtain $c_k$ cluster centroids and then perform the triangle encoding operation from lines 2 to 13. We repeat the lines from 2 to 13 for the extraction of proposed features of testing initial deep features, based on the centroids $\{c_k\}$ learned from training features.

\section{Experimental Results}
This section is divided into three sub-sections: Section \ref{dataset} explains the dataset used; Section \ref{implementation} explains our experimental setup; Section \ref{k_analysis} discusses the analysis of different values of $k$ in the experiment; and Section \ref{analysis} discusses the results and testing time.

\subsection{Dataset}
\label{dataset}
We conduct experiments on the Flickr images sampled from the only available privacy image dataset, PicAlert\cite{zerr2012privacy}, which was provided by Spyromitros et al. \cite{spyromitros2016personalized}.
The dataset contains two categories of images: private and public.
The number of private images in the dataset is lower than public images and we follow the similar configurations as suggested by Tonge et al. \cite{tonge2016image} for the train/test split in the experiment. 
The total number of images is $4700$, in which, $3917$ ($83\%$) images are for training and $783$ ($17\%$) images are for testing. 
Similarly, the ratio of private/public images in each subset (training and testing) is $3:1$.

\subsection{Experimental setup}
\label{implementation}

The experiments have been performed on a laptop with NVIDIA 1050 GeForce GTX GPU and 16GB RAM. We use the keras\cite{chollet2017kerasR} package implemented in R\cite{rcite}, which is open source. Also, we test our proposed unsupervised deep features by utilizing the L2-regularized Logistic Regression (LR) classifier in Liblinear\cite{Fan:2008:LLL:1390681.1442794}.
We fix bias as $1$ and tune $C$, which is the main parameter to tune in L2-regularized Logistic Regression (LR) classifier. 
The grid search technique is used for $C$ in the range of $1$ and $50$, to search the optimal value.
\subsection{Analysis of $k$}
\label{k_analysis}
To select a best $k$, the number of clusters for our dataset, we perform an analysis using the features extracted from the ResNet-50(47) method in the experiment. The tested values for $k$ are in the range of 100 and 500 as seen in Table ~\ref{tab:k}.
While observing in Table~\ref{tab:k}, we notice that the number of cluster $k=250$ yielded a more prominent classification accuracy (\textbf{85.69}\%) than other values. Thus, we empirically employed $250$ as the number of clusters for K-means clustering to extract the proposed unsupervised deep features (UDF).
\begin{table}[tb]
\centering
\caption{Analysis of different $k$, number of clusters, using classification accuracy (\%) while extracting unsupervised deep features (UDF) using ResNet-50(47) method. 
}
\begin{center}
\label{tab:k}
\begin{tabular}{m{2cm} m{1cm} m{1cm} m{1cm}  m{1cm} m{1cm} m{1cm} m{1cm} m{1cm} m{1cm}}
\hline
\textbf{$k$} &100 &150 &200 &250 &300 &350 &400 &450 &500\\
\hline 
\textbf{Accuracy} &84.54 &84.92 &85.05 &\textbf{85.69} &85.18 &85.18 &85.05 &85.18 & 85.56 \\
\hline

\end{tabular}
\setlength{\tabcolsep}{2cm}
\end{center}
\end{table}

\subsection{Analysis of results}
\label{analysis}
We discuss the results of classification accuracy and prediction timings in this section. 

\subsubsection{Classification accuracy}
We compare the proposed features with the state-of-the-art features (deep features extracted from various pre-trained deep learning models), in terms of classification accuracy.
To examine what deep features are more effective, we evaluate the deep features from six different layers of ResNet-50 model.
In Table~\ref{tab:4}, we see that our proposed unsupervised deep features extracted from each layer outperform the existing features of the corresponding layer. The highest accuracy is from the activation layer 48 (ResNet-50(48)), which is \textbf{85.95\%}, among all unsupervised deep features. Similarly, the least accuracy is generated by the ResNet-50(42) which is 84.80\%. We notice the interesting result from the ResNet-50(avg\_pool) layer whose accuracy (85.56\%) is same for both kinds of features. It is a top layer of the ResNet-50 model, which carries important information about objects in the images. 

In spite of a lower size, the classification accuracies of the proposed features are consistently increased for each layer\cite{he2016deep} except the top layer, compared to the corresponding original deep features.
\begin{table}[t]
\centering
\caption{Comparisons of the proposed unsupervised deep features (UDF) with the initial deep features (IDF) with regard to classification accuracy (\%).
}
\begin{center}
\label{tab:4}
\begin{tabular}{m{5cm} m{2.2cm} m{2.2cm}}
\hline
\textbf{Methods} & \textbf{IDF} & \textbf{UDF}\\
\hline 
\textbf{ResNet-50(42)} & 83.90 & \textbf{84.80} \\
\textbf{ResNet-50(44)} & 84.03 &\textbf{85.05}\\
\textbf{ResNet-50(45)}  & 85.05 &\textbf{85.82}\\
\textbf{ResNet-50(47)} & 84.16 &\textbf{85.69}\\
\textbf{ResNet-50(48)} & 84.41 &\textbf{85.95} \\
\textbf{ResNet-50(avg\_pool)}& \textbf{85.56} & \textbf{85.56} \\
\hline
\end{tabular}
\setlength{\tabcolsep}{2cm}
\end{center}
\end{table}
\begin{table*}[t]
\centering
\caption{Comparisons of the proposed features with the state-of-the-art deep features, which are extracted from different pre-trained deep learning models, in terms of classification accuracy (\%) and testing time (seconds). }
\begin{center}
\label{tab:5}
\begin{tabular}{m{6cm} m{2cm} m{1.6cm} m{1.7cm}}
\hline
\textbf{Methods} &\textbf{Feat. size}& \textbf{Acc.} & \textbf{Test. time}\\
\hline
\textbf{VGG-16}(${FC_1}$)\cite{simonyan2014very}&4096-D &84.67 &0.120\\
\textbf{VGG-16(\textbf{${FC_2}$})}\cite{simonyan2014very}&4096-D &84.80&0.090\\
\textbf{VGG-19(${FC_1}$)}\cite{simonyan2014very} &4096-D&84.67&0.060\\
\textbf{VGG-19(${FC_2}$)}\cite{simonyan2014very}&4096-D&84.54&0.090\\
\textbf{Inception-V3(avg\_pool)}\cite{szegedy2016rethinking}&2048-D&74.84&0.050\\
\textbf{DenseNet-121(avg\_pool)}\cite{huang2017densely}&1024-D&79.56&0.025\\
\textbf{DenseNet-169(avg\_pool)}\cite{huang2017densely}&1664-D&78.41&0.030\\
\textbf{DenseNet-201(avg\_pool)}\cite{huang2017densely}&1920-D&79.05&0.020\\
\textbf{Xception(avg\_pool)}\cite{chollet2017xception}&2048-D&75.00&0.050\\
\textbf{Inception-ResNet-v2(avg\_pool)}\cite{szegedy2017inception}& 1536-D&74.96 &0.020\\
\textbf{Ours (Serial Fusion)}&500-D&\textbf{86.33} &\textbf{0.015}  \\
\hline
\end{tabular}
\end{center}
\end{table*}
Furthermore, to improve the classification for privacy images, we fuse two unsupervised deep features. We tested the combination of two different deep features and empirically found that the combination of ResNet-50(47) and ResNet-50 (avg\_pool) produces a higher separability. That is, the resulting features become more discriminable than other types of combinations.
We use the serial feature fusion strategy\cite{yang2003feature} which produces 500-D features in total. 
The comparisons of our fused features with the state-of-the-art deep features are shown in Table~\ref{tab:5}. The compared deep features are extracted from various pre-trained deep learning models: VGG-Net\cite{simonyan2014very} (VGG-16 and VGG-19), ResNet-50\cite{he2016deep}, DenseNet-121\cite{huang2017densely}, DenseNet-169\cite{huang2017densely}, DenseNet-201\cite{huang2017densely}, Inception-V3\cite{szegedy2016rethinking}, Xception\cite{chollet2017xception}, Inception-ResNet-v2\cite{szegedy2017inception}.
We observe that the lowest accuracy is 74.84\% from Inception-ResNet-v2\cite{szegedy2017inception}.  VGG-Net\cite{simonyan2014very} with VGG-16($FC2$) features yield an accuracy of 84.80\%  (which is the second highest accuracy on the dataset), which clearly benefits from a greater feature size.
Our fused deep features produce an accuracy of \textbf{86.33}\% which is 11.49\% higher than the lowest accuracy\cite{szegedy2017inception}. 
The features from other pre-trained models except VGG-Net\cite{simonyan2014very} and ResNet-50\cite{he2016deep} are not appropriate for the classification of privacy images because of their lower classification accuracies. We notice that our proposed features outperform the existing features in terms of classification accuracy.

\subsubsection{Testing time}
We also analyze the efficiency of our proposed deep features, i.e., the testing time during classification. The testing time of the proposed unsupervised features is compared with those of the state-of-the-art deep features (Table \ref{tab:5}). 
The testing time is measured in seconds.
Our fused features achieve \textbf{0.015} seconds and is the fastest among all.
We also observe that the testing timings of the proposed features during classification are shorter compared to the corresponding deep features (Table~\ref{tab:2}). The minimum testing time reported is \textbf{0.003} seconds which is the least among all. This attributes to a lower size of the proposed features than the original deep features: a larger feature size often leads to a slower prediction speed. 
We list the feature sizes of original deep features and the proposed features in Table~\ref{tab:3}. Since we set $250$ as the number of cluster centroids ($k$) during K-means clustering, the size of the proposed features is 250.
Here, we notice that our proposed features outperform state-of-the-art deep features in terms of testing time as well.
\begin{table}[t]
\centering
\caption{Testing timings (in seconds) of the proposed unsupervised deep features (UDF) as well as the initial deep features (IDF).}
\begin{center}
\label{tab:2}
\begin{tabular}{m{4cm} m{3cm} m{2cm}}
\hline
\textbf{Methods} & \textbf{IDF} & \textbf{UDF}\\
\hline 
\textbf{ResNet-50(42)} & 0.017 & \textbf{0.011} \\
\textbf{ResNet-50(44)} & 0.009 &\textbf{0.004}\\
\textbf{ResNet-50(45)}  & 0.015 &\textbf{0.003}\\
\textbf{ResNet-50(47)} & 0.011 &\textbf{0.003}\\
\textbf{ResNet-50(48)} & 0.009 &\textbf{0.003} \\
\textbf{ResNet-50(avg\_pool)} & 0.160 & \textbf{0.003} \\
\hline
\end{tabular}
\end{center}
\end{table}
\begin{table}[t]
\centering
\caption{Sizes of the proposed unsupervised deep features (UDF) and the initial deep features (IDF).}
\begin{center}
\label{tab:3}
\begin{tabular}{m{4cm} m{3cm} m{3cm}}
\hline
\textbf{Methods} & \textbf{IDF} & \textbf{UDF}\\
\hline 
 \textbf{ResNet-50(42)} & 512-D & 250-D \\
\textbf{ResNet-50(44)} & 512-D &250-D\\
\textbf{ResNet-50(45)}  & 512-D &250-D\\
\textbf{ResNet-50(47)} & 512-D &250-D\\
\textbf{ResNet-50(48)} & 512-D &250-D \\
\textbf{ResNet-50(avg\_pool)} & 2048-D & 250-D \\
\hline
\end{tabular}
\end{center}
\end{table}

\section{Conclusion}

In this paper, we have introduced the unsupervised deep features based on the deep features extracted from the ResNet-50 model. 
We first extract the deep features from top activation layers of the ResNet-50 model for each image, and then perform the K-means clustering over training set to learn the centroids. Finally, we encode the computed features to a feature vector for each image based on the learned centroids. The feature vector is taken as an input to our trained model which gives the prediction. Experiments show that our proposed features are more accurate in privacy image classification and produce shorter testing time than state-of-the-art deep features. In the future, we would like to investigate a more complicated classification of privacy images which involve more than two categories.

\bibliographystyle{splncs04}
\bibliography{psivt_reference}

\end{document}